\pdfoutput=1

\documentclass[11pt]{article}
 \usepackage{multirow}
 \usepackage{hyperref}
\usepackage[preprint]{coling}

\usepackage{times}
\usepackage{latexsym}
\usepackage{longtable}

\usepackage{colortbl}
\usepackage{xcolor}
\usepackage[T1]{fontenc}
\usepackage{amsmath}
\usepackage{graphicx}
\usepackage{hyperref}
\usepackage[utf8]{inputenc}

\usepackage{microtype}

\usepackage{inconsolata}

\usepackage{graphicx}

\title{Instructions for COLING 2025 Proceedings}

\author{Zhe Ren \\
  \texttt{renzhe@stu.xju.edu.cn} \\}

\usepackage{amssymb}

\usepackage{hyperref}

\begin{document}
\title{Prompt Tuning for Few-Shot Continual Learning Named Entity Recognition}
\maketitle

\begin{abstract}
Knowledge distillation has been successfully applied to Continual Learning Named Entity Recognition (CLNER) tasks, by using a teacher model trained on old-class data to distill old-class entities present in new-class data as a form of regularization, thereby avoiding catastrophic forgetting. However, in Few-Shot CLNER (FS-CLNER) tasks, the scarcity of new-class entities makes it difficult for the trained model to generalize during inference. More critically, the lack of old-class entity information hinders the distillation of old knowledge, causing the model to fall into what we refer to as the \textit{ Few-Shot Distillation Dilemma}. In this work, we address the above challenges through a prompt tuning paradigm and memory demonstration template strategy. Specifically, we designed an expandable \textbf{A}nchor words-oriented \textbf{P}rompt \textbf{T}uning (APT) paradigm to bridge the gap between pre-training and fine-tuning, thereby enhancing performance in few-shot scenarios. Additionally, we incorporated \textbf{M}emory \textbf{D}emonstration \textbf{T}emplates (MDT) into each training instance to provide replay samples from previous tasks, which not only avoids the  \textit{Few-Shot Distillation Dilemma} but also promotes in-context learning. Experiments show that our approach achieves competitive performances on FS-CLNER.
\end{abstract}
\begin{figure}[htbp]
\centerline{\includegraphics [width=0.5\textwidth]{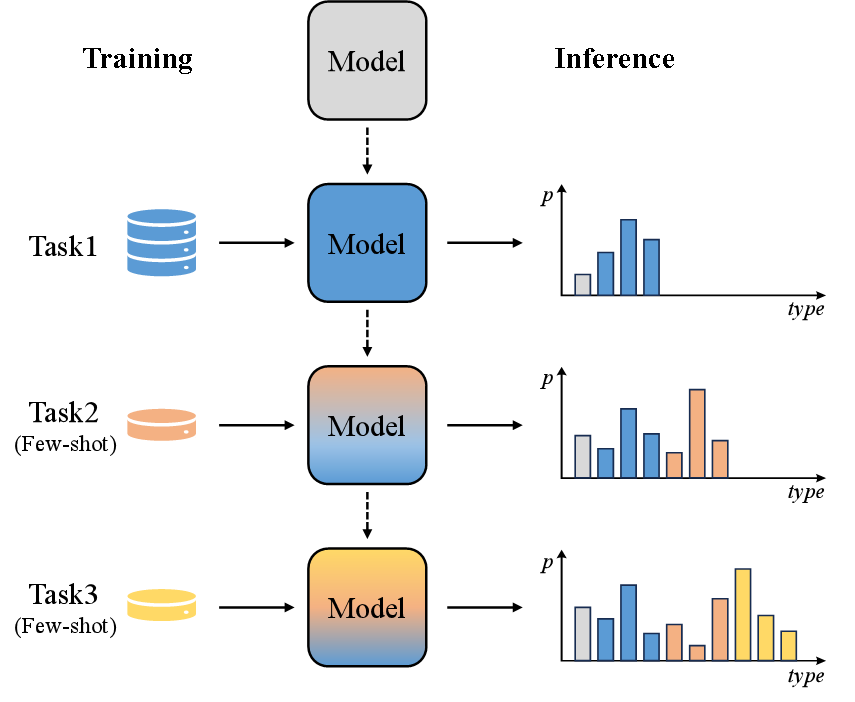}}
\caption{An illustration of the FS-CLNER task.}
\label{fig.1}
 
\end{figure}

\section{Introduction}

Named Entity Recognition (NER) plays a crucial role in the practical application of natural language processing (NLP). Traditional NER models are typically trained on large-scale datasets with predefined entity types and then deployed to extract these entities from unstructured text data without further adjustment or refinement. However, in many real-world scenarios, new entity types may emerge periodically, and available training data for these new entities is often scarce. While a natural yet inelegant solution would be to retrain the model by adding new class data to the original old class data, this approach may be infeasible due to privacy concerns or memory limitations \cite{ma2020adversarial}. Therefore, an ideal NER model should be able to learn these new entities (i.e., plasticity) from minimal data without compromising its existing capabilities (i.e., stability) to meet dynamic demands. This, however, poses a significant challenge for traditional NER models.
\par To enable NER models to adapt to dynamic data streams, researchers have explored Continual Learning NER (CLNER) and have made significant progress. Mainstream approaches are based on knowledge distillation (\citealp{monaikul2021continual},  \citealp{zhang2023neural}), where the core idea is to use a teacher model trained on old-class data to distill old-class entities found in new-class data as a form of regularization, allowing the model to learn new-class entities without forgetting old-class entities. However, when annotated data for new classes is scarce, existing CLNER methods face two major challenges: (1) the limited information on new-class entities in the sparse training data results in poor generalization of the trained model during inference; (2) the new-class training data contain almost no old-class entity information, which obstructs the distillation of old knowledge and leads to catastrophic forgetting, a phenomenon we refer to as the  \textit{Few-Shot Distillation Dilemma}. These issues have spurred research into more challenging Few-Shot CLNER (FS-CLNER), as shown in \autoref{fig.1}. \citet{wang2022few} conducted the first study on this task, proposing a method that follows the knowledge distillation framework by generating synthetic data of old classes through model inversion, serving as replay data for old entity classes. However, the process of generating synthetic data is complex and time-consuming, requiring careful design of adversarial matching to ensure the effectiveness and authenticity of the synthetic data.

\begin{figure}[htbp]
\centerline{\includegraphics [width=0.5\textwidth]{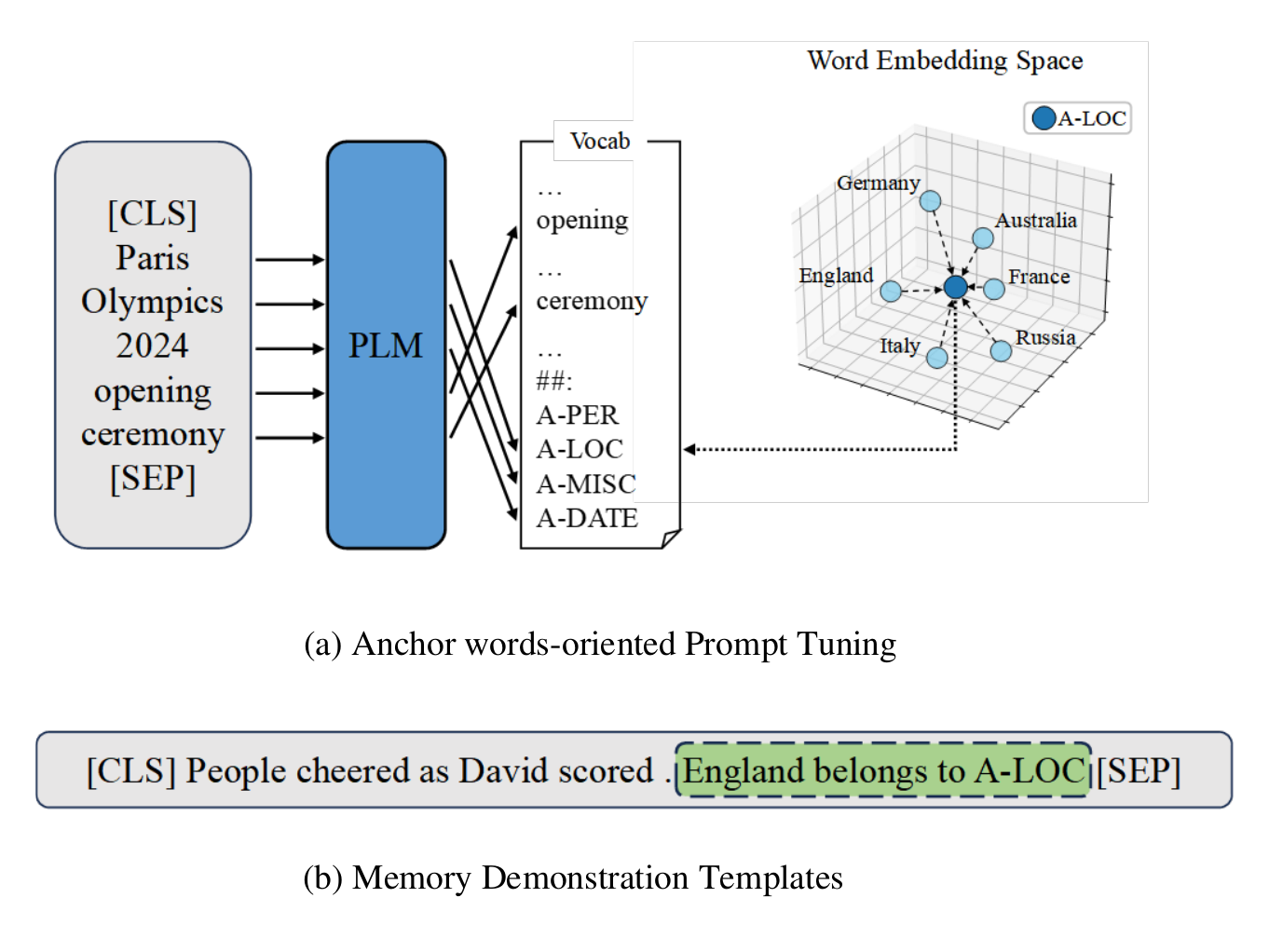}}
\caption{We enhance the model's generalization in few-shot scenarios with an expandable anchor words-oriented prompt tuning paradigm and effectively avoid the few-shot distillation dilemma using memory demonstration templates.}
\label{fig2}
\end{figure}

In this work, we propose a simple and efficient method to address the challenges in FS-CLNER. Inspired by prompt-based NER methods , we redesign the NER task into an expandable 
\textbf{A}nchor words-oriented \textbf{P}rompt \textbf{T}uning (APT)
 paradigm. In this paradigm, the NER classification task is reformulated as a language modeling task, allowing the language model to predict entity mentions as corresponding anchor words. Anchor words are virtual tokens created by merging several representative entity words of the same type, which dynamically expand according to the task flow, as illustrated in  \autoref{fig2} (a). This design narrows the gap between pre-training and fine-tuning caused by differing training objectives, thereby enhancing generalization performance in few-shot scenarios \cite{gao2020making}. Additionally, we incorporate 
 \textbf{M}emory \textbf{D}emonstration \textbf{T}emplates (MDT)
  into each training instance, as shown in \autoref{fig2}(b). These demonstration templates not only act as replay samples for old entities, effectively addressing the Few-Shot Distillation Dilemma, but also complement the expandable Anchor Words-oriented prompt tuning paradigm, enhancing the flow of information in context learning and guiding the language model to better understand the task \cite{wang2023label}.
Our proposed method collaborates with knowledge distillation in a manner similar to ExtendNER \cite{monaikul2021continual}, but differs in that our approach does not require extending the classification head to accommodate new entity types. Rather, it achieves adaptability by dynamically extending the vocabulary with anchor words representing new entity types. Results from experiments on the CoNLL2003 \cite{sang2003introduction} and Ontonote 5.0 \cite{zhao2019gender} datasets under 5-shot and 10-shot FS-CLNER settings show that our method achieves competitive performance without the need for any additional data (such as complex synthetic data), demonstrating its superiority and practical value.
The contributions of this work are summarized as follows:
\par • We successfully introduced prompt tuning to the FS-CLNER task, providing a new perspective on the task.
\par • By using memory demonstration templates, we effectively avoided the  \textit{ Few-Shot Distillation Dilemma}, enhancing the model's adaptability to few-shot dynamic data streams.
\par • Experiments demonstrate that our method does not require additional data (such as complex synthetic data) for FS-CLNER tasks, showcasing its practicality and effectiveness.



 \section{Related Work}

 \textbf{Continual learning.} Human continual learning, also known as lifelong learning, refers to an individual's ability to continuously acquire and adapt to new knowledge throughout their lifetime without forgetting or interfering with existing knowledge, thereby adapting to an ever-changing world. This concept provides important insights for the development of artificial intelligence (AI), guiding AI systems to better adapt to the complex and dynamic real world (\citealp{chen2022lifelong}; \citealp{parisi2019continual}). However, continual learning faces the well-known challenge of catastrophic forgetting (\citealp{mccloskey1989catastrophic};\citealt{robins1995catastrophic} ;\citealp{goodfellow2013empirical} ; \citealp{kirkpatrick2017overcoming}), as neural networks typically update all network parameters via backpropagation when training on new tasks, leading to a sharp decline in the performance of old tasks after learning new ones \cite{de2021continual}. As a result, a range of studies has emerged to explore ways to overcome catastrophic forgetting.
\par Early research on CL primarily focused on image classification tasks in Computer Vision (CV). \citet{li2017learning} introduced the Learning without Forgetting method, which integrates the knowledge distillation framework.\citet{wang2022learning}  proposed a prompt-based CL framework, L2P, to address challenges in CL. These methods were later extended to sentence-level CL tasks in NLP. \citet{sun2020distill} applied DnR distillation and replay to text classification tasks, and \citet{zhu2022continual} applied the prompt-based CL framework to dialogue state tracking tasks. However, these methods are difficult to directly apply to token-level CL tasks, such as CLNER. Currently, mainstream CLNER methods are based on knowledge distillation.\citet{monaikul2021continual} were the first to adopt the knowledge distillation framework for CLNER, while \citet{xia2022learn}added a rehearsal stage, using synthetic samples of old classes to augment the dataset. \citet{zhang2023neural} improved upon this with a span-based CLNER model.
\par Unfortunately, these CLNER methods perform poorly in few-shot settings, facing challenges related to few-shot generalization and the distillation dilemma. \citet{wang2022few} were the first to explore FS-CLNER, proposing a method similar to L\&R \citet{xia2022learn}, which generates synthetic data for old classes to avoid the few-shot distillation dilemma. However, the process of constructing synthetic data is complex and time-consuming, requiring careful design of adversarial matching to ensure the validity and authenticity of the synthetic data. In contrast, our method achieves comparable performance without the need for synthetic data.

\par \textbf{Prompt-based Few-Shot Learning.} The goal of few-shot learning is to emulate the human ability to learn from a small number of examples. In contrast to traditional supervised learning, which requires large amounts of data, few-shot learning relies on only a few labeled examples to make accurate predictions, significantly reducing the time and financial costs associated with data annotation.

The release of GPT-3 \cite{brown2020language}sparked significant interest in prompt-based learning. Unlike traditional fine-tuning methods, where the output layer of a pre-trained model is replaced and fine-tuned for downstream tasks, prompt-based tuning reformulates downstream tasks to align with the format of pre-training, thereby narrowing the objective gap between pre-training and fine-tuning and fully leveraging the potential of pre-trained language models (PLM). As a result, even with limited training samples, PLM can adapt to downstream tasks more quickly ,\citet{schick2020exploiting} were the first to introduce prompt templates into the NER task, demonstrating superior performance in few-shot settings compared to traditional sequence labeling baselines. \citet{ma2021template} later proposed a template-free approach while maintaining the prompt tuning paradigm. \citet{shen2023promptner} unified entity recognition and classification in NER through dual-slot multi-prompt templates. However, these prompt-based few-shot NER methods are not designed to handle dynamic data streams. To the best of our knowledge, we are the first to introduce prompt tuning to FS-CLNER.









\begin{figure*}[ht]
    \centering
      \includegraphics[scale=0.4]{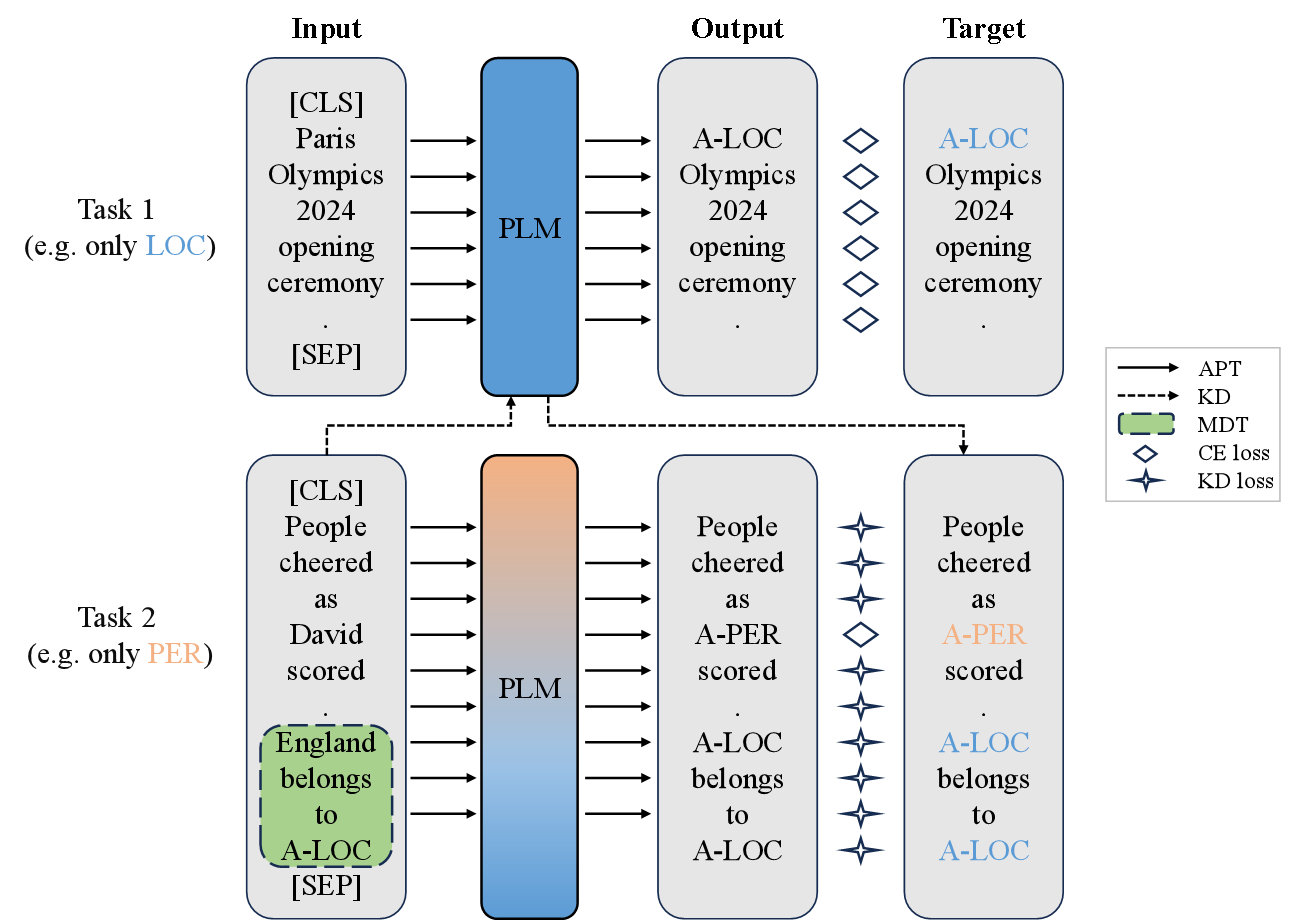}
    
\caption{The overall structure of our proposed method. Solid arrows represent anchor words-oriented prompt tuning, dashed arrows denote knowledge distillation, green areas indicate memory demonstration templates, diamonds signify cross-entropy loss with the target, and stars represent KL divergence with knowledge distillation logits.}
    \label{fig3}
\end{figure*}

\section{Method}
\subsection{Problem Formalization}

 Assume there is a continuous sequence of tasks %
$\{ 1, \cdots ,T\} $, corresponding to the sequence of NER training datasets 
$\{ {{\cal D}^1}, \cdots ,{\cal D}^{T\}}\} $ is a base dataset with a large amount of data, and ${{\cal D}^1}$ are few-shot datasets. If  $|{\cal D}|$ represents the size of a dataset, then 
$\forall t > 1 \Rightarrow  |{{\cal D}^t}| \ll |{{\cal D}^1}|$. Each ${{\cal D}^t} = \{ (X_i^t,Y_i^t)\} _{i = 1}^{|{{\cal D}^t}|}$, where 
$X_i^t = [x_i^{t,1}, \cdots ,x_i^{t,{N_i}}]$ and 
$Y_i^t = [y_i^{t,1}, \cdots ,y_i^{t,{N_i}}]$ represent the token sequences and label sequences of length ${N_i}$, respectively. The entity type set contained in ${{\cal D}^t}$  is ${E^t} = \{ e_t^i\} _{i = 1}^{{c_t}}$, where 
${c_t}$ is the number of entity types in the -th task. The entity types in different tasks do not overlap, i.e., $\forall i,j \in \{ 1, \cdots ,{\rm T}\} , i \ne j \Rightarrow  {E^i} \cap {E^j} = \emptyset $. The goal of few-shot CLNER is to sequentially train on different tasks, and after the -th task, the model should be able to infer and recognize all previously seen entity types 
$\{ {E^i}\} _{i = 1}^t$ .
\par Since  ${\{ {{\cal D}^t}\} _{t > 1}}$ is a few-shot dataset, models trained on these data exhibit weak generalization ability during inference. To address this, we designed a prompt tuning paradigm oriented toward anchor words, as described in   \hyperref[sec:3.2]{3.2}  . Moreover, %
${\{ {{\cal D}^t}\} _{t > 1}}$ contains little to no old entity type information, which leads the model to fall into the "few-shot distillation dilemma." To tackle this issue, we proposed a memory demonstration template strategy to augment each  in %
$\{ X_i^t\} _{i = 1}^{|{{\cal D}^t}|}$ , as detailed in \hyperref[sec:3.3]{3.3}. The overall architecture is illustrated in \autoref{fig3}.
\subsection{Anchor Words-oriented Prompt Tuning}
\label{sec:3.2}

Inspired by the \cite{ma2021template} , we adopt an expandable anchor words-oriented prompt tuning paradigm to address the issue of poor generalization in few-shot scenarios. Unlike \cite{ma2021template}, we also account for incremental settings by dynamically expanding the anchor words.

Formally, taking the $t$-th task as an example, we first construct the anchor word set ${A^t}$ for the current task entity type set . Specifically, for each type $e_t^n(1 \le n \le {c_t})$ , we select the top $K$ entities that best represent that class to form the entity word set $\xi _t^n$ . The anchor word for this class is represented by  ${\cal A}(e_t^n)$ (such as $A{\rm{ - }}LOC$ ), where ${\cal A}:E \to A$ is the mapping function that maps the entity type to a virtual anchor word. At this point, the embedding vector for the anchor word is defined as:
\begin{equation}
\mathbb {E}\left( \mathcal{A}\left(e_{t}^{n}\right) \right) = \frac{1}{K} \sum_{\varepsilon \in \xi_{t}^{n}} \mathbb{E}(\varepsilon)
\end{equation}

Where ${\mathbb E}( * )$ represents the word embedding from the PLM. Suppose there is an input sequence $X_i^t = [x_i^{t,1}, \cdots ,x_i^{t,j}, \cdots ,x_i^{t,{N_i}}]$ , where the label of $x_i^{t,j}$ is $e_t^n$ and the rest are classified as type ${\rm O}$. We build the target sequence $\tilde X_i^t = [x_i^{t,1}, \cdots ,{\cal A}(e_t^n), \cdots ,x_i^{t,{N_i}}]$ by replacing $x_i^{t,j}$ with the corresponding anchor word. During training, the word embeddings of the input sequence $X_i^t$ are first fed into a BERT \cite{devlin2018bert} encoder to obtain the contextual embeddings:
\begin{equation}
    H_i^t = {\rm{Encoder}}({\rm E}(X_i^t))
\end{equation}

Here, 
$H_i^t \in {{\mathbb R}^{{N_i} \times {d^h}}}$ is the representation from the encoder's hidden layer, where ${d^h}$ is the size of the hidden layer. Unlike traditional sequence labeling tasks, we do not introduce a new classification head; instead, we use the original MLM head to predict the probability distribution:

\begin{equation}
 z_i^{t,j} = {W_{MLM}}h_i^{t,j} + {b_{MLM}} 
\end{equation}

\begin{equation}
\begin{split}
 P(x_i^{t,j} = \tilde{x}_i^{t,j} | X_i^t) 
 &= \mathrm{softmax}(z_i^{t,j}) \\
 &= \frac{\exp(z_i^{t,j})}
 {\sum\limits_{v \in (\mathcal{V} \cup A^t)} \exp(z_{i,v}^{t,j})}
\end{split}
\end{equation}

Where ${W_{MLM}}$ and ${b_{MLM}}$ are the weights and biases of the MLM head, $z_i^{t,j}$ is the logits vector corresponding to $x_i^{t,j}$ , and  ${\cal V}$ is the original vocabulary of the model. As no new parameters are introduced, the model is easier to adapt to target tasks with fewer samples. Ultimately, the model is optimized through cross-entropy loss:

\begin{equation}
\begin{aligned}
    \mathcal{L}_{PT} = 
    - \frac{1}{N_i} \sum\limits_{n = 1}^{N_i} 
    &\sum\limits_{m = 1}^{|\mathcal{V}| + c_t} 
    1(\tilde{x}_i^{t,n} = m) \\
    &\times \log P(\tilde{x}_i^{t,n} = m | X_i^t)
\end{aligned}
\end{equation}

Here, $1(\tilde x_i^{t,n} = m)$ is an indicator function that takes the value 1 when the target label of the $n$-th token is $m$ , and 0 otherwise. During the inference process, only one decoding step is required to obtain all the labels of the input sequence:
\begin{equation}
    P(y_i^{t,j} = e_t^n|X_i^t) = P(x_i^{t,j} = {\cal A}(e_t^n)|X_i^t)
\end{equation}

Overall, the expandable anchor word-oriented prompt tuning has two advantages: 1) It does not require a specified template and only needs a single decoding step; 
2) It maintains the pre-training paradigm, fully utilizing the potential of the PLM and improving the few-shot learning capability.
\subsection{Memory Demonstration Template}
\label{sec:3.3}
In the FS-CLNER task, the few-shot training data in the $t$-th task stage contains almost no information about old class entities $\{ {E^i}\} _{i = 1}^{t - 1}$ , making it impossible to transfer this old knowledge to the current stage through distillation, leading to a few-shot distillation dilemma. To address this challenge, we have set up a memory demonstration template strategy. Specifically, we add automatically created memory demonstration templates to each piece of training data in the current stage, providing replay examples for distillation and inputting them into the LM. The format of the memory demonstration templates adopts an entity-oriented demonstration approach, which is consistent with prompt tuning and complements it effectively.

Formally, for the $t$-th task stage, assuming $e_i^n(1 < i < t, 1 < n < {c_i})$ is one of the old class entities to be distilled, we define the format of the memory demonstration template ${\cal T}$ as "$[Entity]$  belongs  to $ [ANCHOR].$". In this format, the first slot is randomly filled with  $\varepsilon (\varepsilon  \in {\xi _n})$ , providing old class entity information for the input sequence; the second slot is filled with the corresponding anchor word ${\cal A}(e_t^n)$ , which complements the prompt tuning goal oriented towards scalable anchor words. The format of the corresponding template target sequence  $\tilde {\cal T}$is "${\cal A}(e_t^n)$ belongs to ${\cal A}(e_t^n).$". For example, the memory demonstration template for the entity type LOC is "$England$  belongs to $A{\rm{ - }}LOC.$", and the corresponding target sequence is "$A{\rm{ - }} LOC$ belongs to  $A{\rm{ - }}LOC.$". Subsequently, the input sequence and its target sequence are expanded as:
\begin{equation}
    (X'_i)^{t} = [X_i^{t}, \mathcal{T}], \quad (\tilde{X}'_i)^{t} = [\tilde{X}_i^{t}, \tilde{\mathcal{T}}]
\end{equation}

Similarly, after adding multiple memory demonstration templates corresponding to old-class entities to the input sequence, each input will contain comprehensive and diverse old-class entity information. Note that no memory demonstration template is added during the inference process.
\par In summary, memory demonstration templates have two benefits: 1) They provide replay examples about past memories, helping to overcome the few-shot distillation dilemma and prevent catastrophic forgetting; 2) Through entity-oriented demonstration examples for anchor words, they further clarify the goal of prompt tuning and enhance the flow of information in the context, guiding the language model to better understand the task \cite{wang2023label}.

\subsection{Knowledge Distillation}
Our proposed method generally follows the knowledge distillation-based CLNER framework, as shown in \autoref{fig3}. First, we feed the current task's data into the teacher model for forward propagation (indicated by the dashed arrows in the figure), using the results as pseudo-labels to jointly train the student model with the current task's gold labels. Unlike previous work \cite{monaikul2021continual}, our method does not need to expand the output layer to accommodate new entity types, but instead dynamically extends anchor words to fit new entity types.
\par Formally, suppose the model trained in the task $t - 1  $ is ${{\rm M}^{t - 1}}$, and the model has learned $\sum\nolimits_{i = 1}^{t - 1} {{c_i}} $ entities, with its output dimension being $|{\cal V}| + \sum\nolimits_{i = 1}^{t - 1} {{c_i}} $. In the current task  $t$ stage, assuming $x_i^{t,j}$ is the entity of the current task, we first use model ${{\rm M}^{t - 1}}$ to predict the extended $(X'_i)^{t}$ , taking the logits values at all positions except those of the current task's gold entities as pseudo-labels. These are used to jointly train the student model ${{\rm M}^t}$ with the gold entity labels. For the pseudo-label part, we aim to minimize the KL divergence between the student's output distribution and the teacher's output distribution to optimize the model to learn old knowledge (as shown in the diamond part of the figure):

\begin{equation}
\begin{aligned}
{{\cal L}_{KD}} &= \frac{1}{{N_i}} \sum_{n = 1}^{N_i} \sum_{m = 1}^{|{\cal V}| + {c_t}} \\
&\quad P_{{\rm M}^{t - 1}}(\tilde{x}_i^{t,n} = m) \\
&\quad \times \log \left( \frac{P_{{\rm M}^{t - 1}}(\tilde{x}_i^{t,n} = m)}{P_{{\rm M}^t}(\tilde{x}_i^{t,n} = m)} \right)
\end{aligned}
\end{equation}

For the gold label part , we use ${{\cal L}_{PT}}$ to encourage learning new knowledge (as shown in the star-shaped part of the figure). Ultimately, the model's total loss is composed of the following two parts:

\begin{equation}
   {{\cal L}_{tot}} = \alpha {{\cal L}_{KD}} + \beta {{\cal L}_{PT}}
\end{equation}Where $\alpha $ and $\beta $ are weighting factors.

\section{Experimental Settings}
\label{sec:bibtex}
\subsection{Datasets}

Following the previous FS-CLNER work, we use CoNLL2003 and Ontonote 5.0 as the original datasets, respectively, and construct FS-CL datasets by reorganizing the original data. Each FS-CL dataset is divided into a base class stage (task 1) and incremental stages (subsequent tasks). The training data for the base class stage comes from the original training set, while the training data for the incremental stages is sampled from the original validation set. The test set for all stages uses the original test set. Notably, we do not set a validation set, as this better aligns with the practical requirements of few-shot scenarios in real-world applications.



\subsection{FS Settings}

The training data for the incremental stages is obtained through greedy sampling \cite{yang2020simple} from the original validation set. We conducted 5-shot and 10-shot on CoNLL2003 and 5-shot on Ontonotes 5.0. For detailed experimental settings, please refer to the Appendix.
\subsection{CL Settings}
The task divisions and different orderings for both datasets strictly follow previous work, as detailed in Tables 3 and 4 in the Appendix. Additionally, to ensure fairness and avoid biases that may arise from different reorganization strategies (detailed in the Appendix), we follow the approach of the SpanKL \cite{zhang2023neural}, performing a thorough evaluation of all possible reorganization strategies.

\begin{figure*}[ht]
    \centering
     \includegraphics[width=1\linewidth]{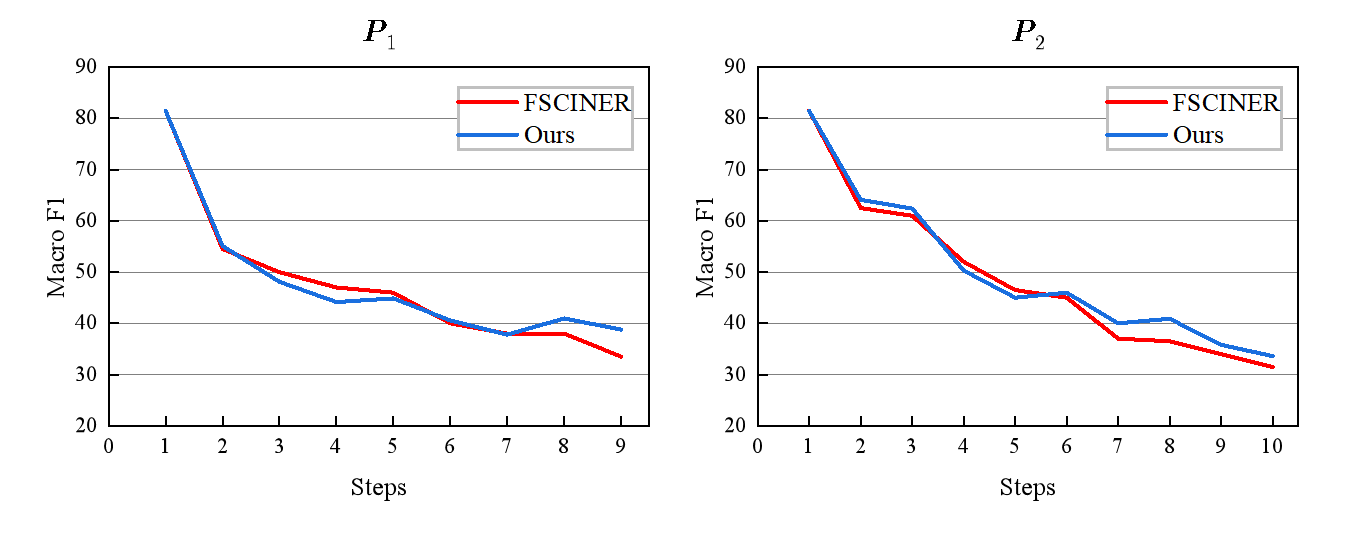}

\caption{Results on OntoNote 5.0 for two permutations in the 5-shot CL setting. Since the original results of the baseline were presented in a line chart without specific numerical values, we estimated the data points by visually interpreting the chart. Although we took care to minimize potential errors, this estimation might introduce slight discrepancies in the exact values.}
    \label{fig4}
\end{figure*}

\begin{table*}[]
\caption{ Results on CoNLL2003. † denotes official reported results, and * indicates corrections made to the official results. The\colorbox[HTML]{8ED873}{\textbf{best}} and \colorbox[HTML]{D9F2D0}{\textbf{second best}} results have been highlighted.}
\begin{tabular}{cccccccccccc}

\hline
\multicolumn{1}{l}{\textbf{}} & \multicolumn{5}{c}{\textbf{5-shot}} & \textbf{} & \multicolumn{5}{c}{\textbf{10-shot}} \\ \cline{2-6} \cline{8-12} 
 & Step1 & Step2 & Step3 & Step4 & Avg$ \ge $2 & \textbf{} & Step1 & Step2 & Step3 & Step4 & Avg$ \ge $2 \\ \hline
ExtendNER & 88.42 & 44.28 & 37.10 & 36.18 & 39.19 &  & 88.42 & 53.77 & 39.06 & 35.88 & 42.90 \\
AddNER & 88.58 & 47.62 & 38.94 & 38.21 & 41.59 &  & 88.58 & 52.14 & 42.70 & 40.64 & 45.16 \\
SpanKL & 88.59 & 47.51 & 40.14 & 38.66 & 42.10 &  & 88.59 & 52.21 & 43.66 & 40.37 & 45.41 \\ \hline
DTPF† & 87.75& 63.73 & 60.04 & \cellcolor[HTML]{D9F2D0}\textbf{60.30} & 61.36 &  & 87.75 & 68.27 & \cellcolor[HTML]{D9F2D0}\textbf{65.55} & \cellcolor[HTML]{D9F2D0}\textbf{64.55} & \cellcolor[HTML]{D9F2D0}\textbf{66.12} \\
FSCINER† & 88.35 & \cellcolor[HTML]{8ED873}\textbf{71.31} & \cellcolor[HTML]{D9F2D0}\textbf{63.76} & 59.37 & \cellcolor[HTML]{D9F2D0}\textbf{64.81*} &  & 88.35 & \cellcolor[HTML]{8ED873}\textbf{70.75} & 64.60 & 60.02 & 65.12 \\ \hline
\textbf{Ours} & 88.89 & \cellcolor[HTML]{D9F2D0}\textbf{68.21} & \cellcolor[HTML]{8ED873}\textbf{64.96} & \cellcolor[HTML]{8ED873}\textbf{63.54} & \cellcolor[HTML]{8ED873}\textbf{65.57} &  & 88.89 & \cellcolor[HTML]{D9F2D0}\textbf{70.03} & \cellcolor[HTML]{8ED873}\textbf{66.37} & \cellcolor[HTML]{8ED873}\textbf{64.88} & \cellcolor[HTML]{8ED873}\textbf{67.09} \\ \hline
\end{tabular}
\label{table1}
\end{table*}

\begin{table*}[h
]
\caption{The results of our method with different reorganization strategies on CoNLL2003.}
\begin{tabular}{ccccccccccccc}
\hline
 &  & \multicolumn{5}{c}{\textbf{ToA}} &  & \multicolumn{5}{c}{\textbf{ToF}} \\ \cline{3-13} 
 &  & Step1 & Step2 & Step3 & Step4 & Avg$ \ge $2 &  & Step1 & Step2 & Step3 & Step4 & Avg$ \ge $2 \\ \hline
\multirow{2}{*}{\textbf{EoA}} & \textbf{5-shot} & 88.89 & 68.21 & 64.96 & 63.54 & 65.57 &  & 74.79 & 62.57 & 59.92 & 59.10 & 60.53 \\
 & \textbf{10-shot} & 88.89 & 70.03 & 66.37 & 64.88 & 67.09 &  & 74.79 & 65.08 & 62.33 & 61.72 & 63.04 \\ \cline{2-13} 
\multicolumn{1}{r}{\multirow{2}{*}{\textbf{EoF}}} & \textbf{5-shot} & 90.68 & 73.69 & 67.73 & 65.41 & 68.94 &  & 92.01 & 74.10 & 65.95 & 61.30 & 67.12 \\
\multicolumn{1}{r}{} & \textbf{10-shot} & 90.68 & 75.10 & 71.84 & 65.28 & 70.74 &  & 92.01 & 76.59 & 65.22 & 62.34 & 68.05 \\ \hline
\end{tabular}
\label{table2}
\end{table*}

\subsection{Baseline}
 We compared two state-of-the-art FS-CLNER models: \textbf{FSCINER }\cite{wang2022few}, which generates synthetic data through an inverted NER model to address the few-shot distillation challenge, and \textbf{DTPF} \cite{chen2023decoupled}, a decoupled two-stage pipeline framework for FS-CLNER. Additionally, we compared three state-of-the-art CLNER models:  \textbf{AddNER}\cite{monaikul2021continual}, the earliest approach to solving the CLNER problem by adding new classifiers to adapt to new entity types; \textbf{ExtendNER} \cite{monaikul2021continual} , which adapts to new entity types by expanding the dimensions of the old classifier; and \textbf{SpanKL} \cite{zhang2023neural}, a span-based CLNER baseline model. All baseline models use bert-base-cased as the encoder. For models with open-source code, we reproduced their results for comparison; for those without open-source code, we used the results reported in their official publications for comparison.

\label{sec:Main Results}
\section{Main Results}
\subsection{Comparison with Baseline}
\autoref{table1} presents the comparison between our method and baseline models on CoNLL2003. The results show that our proposed method performs exceptionally well in the FS-CLNER task, typically ranking first or second. In few-shot settings, the four conventional CLNER models perform poorly, consistent with our analysis of FS-CLNER: when information about previously learned entity classes is extremely limited, distillation of knowledge from old classes is hindered, leading to the few-shot distillation dilemma. Additionally, the two CLNER models specifically designed for few-shot scenarios perform relatively better at mitigating this issue, but their performance still lags behind our method as task stages increase. Our method demonstrates a significant advantage in later stages. It should be noted that DTPF’s results were obtained under the  setting, and their approach uses K-example sampling rather than strict K-shot sampling. Furthermore, our method does not use CRF decoding, yet it remains competitive.
\autoref{fig4} shows our 5-shot CL results compared to FSCINER on OntoNote 5.0. This CL setting, featuring multiple tasks with potentially more than one entity type per task, poses significant challenges for FS-CL. Despite this, our method outperforms FSCINER in both early and later steps.
\subsection{ Results of Different Reorganization}

To fully evaluate the FS-CL capability of the models, we also report results under different reorganization strategies, as shown in \autoref{table2}. Overall, the results under the  $* \to EoF$ strategy are significantly better than those under $* \to EoA$ , as the latter includes unseen entity types, making it more challenging. Additionally, the $TOA \to *$  strategy generally performs better than the $TOF \to *$ strategy, indicating that negative samples play a positive role in training.

\begin{table*}[]
\caption{We conducted ablation studies on the CoNLL2003 dataset under the 5-shot and 10-shot CL settings. w/o APT indicates the exclusion of Anchor words-oriented Prompt Tuning, and w/o MDT indicates the removal of Memory Demonstration Templates.}
\begin{tabular}{lccccccccccc}
\hline
\multicolumn{1}{c}{} & \multicolumn{5}{c}{\textbf{5-shot}} &  & \multicolumn{5}{c}{\textbf{10-shot}} \\ \cline{2-12} 
\multicolumn{1}{c}{} & Step1 & Step2 & Step3 & Step4 & Avg$ \ge $2 &  & Step1 & Step2 & Step3 & Step4 & Avg$ \ge $2 \\ \hline
\textbf{Ours} & \textbf{88.89} & \textbf{68.21} & \textbf{64.96} & \textbf{63.54} & \textbf{65.57} &  & \textbf{88.89} & \textbf{70.03} & \textbf{66.37} & \textbf{64.88} & \textbf{67.09} \\ \hline
\textit{w/o APT} & 87.72 & 58.41 & 54.33 & 46.20 & 52.98 &  & 87.72 & 60.95 & 55.68 & 48.92 & 55.18 \\
\textit{w/o MDT} & 88.71 & 64.14 & 52.73 & 47.60 & 54.82 &  & 88.71 & 68.03 & 58.83 & 55.80 & 60.89 \\ \hline
\end{tabular}
\label{table3}
\end{table*}

\section{Analysis}
\subsection{The Impact of APT}
To investigate whether APT enhances the model's generalization ability in few-shot settings, we removed the APT module and reported the results, as shown in \autoref{table3}. In this case, the model had to introduce new classification heads, effectively degrading into a model similar to ExtendNER. The results show that removing APT had little effect on performance in the base class stage, but significantly degraded performance in the incremental stages. This suggests that APT improves the model's generalization ability under few-shot conditions, enhancing its plasticity in such scenarios. It is worth noting that the model without APT is equivalent to ExtendNER+MDT. Compared to using ExtendNER alone, the MDT strategy helps alleviate the few-shot distillation dilemma across any base method. We do not analyze the choice of anchor words, as prior work \cite{ma2021template} has already provided such priors, and our focus is on evaluating FS-CL performance based on these priors.


\begin{figure}[ht]
    \centering
     \includegraphics[width=1\linewidth]{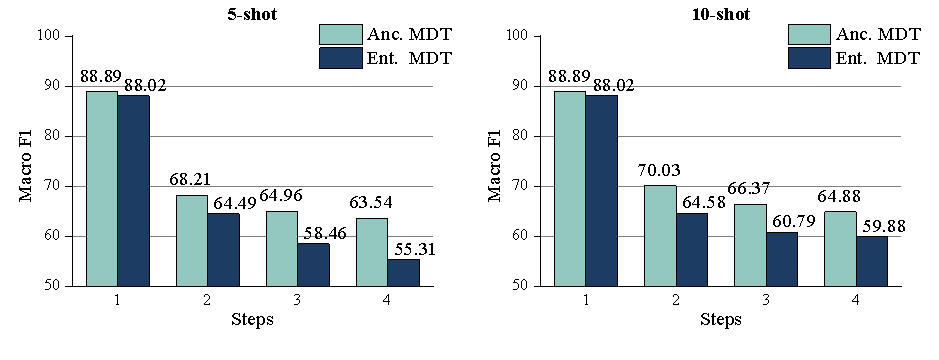}

\caption{Comparison of different formats of MDT. Experiments were conducted on the CoNLL2003 dataset under both 5-shot and 10-shot CL settings.}
    \label{fig5}
\end{figure}

\subsection{The Impact of MDT}

We explored the impact of MDT on CL performance in few-shot scenarios and reported the results after removing MDT, as shown in \autoref{table3}. After removing MDT, the model's performance heavily relied on the strict requirement that the training samples of the current task include entities from previous tasks. When this requirement was not met, the model's performance significantly declined, indicating that MDT effectively mitigates the few-shot distillation dilemma.
We further investigated the impact of different MDT formats on the model's performance, as shown in \autoref{fig5}. Under the premise that MDT can serve as replay samples, we designed the following two formats: Anchor word-oriented MDT(Anc. MDT), with the format , which is the format used in this paper; Entity word MDT(Ent. MDT), with the format , such as "England." The results show that the anchor word-oriented template provides clearer category guidance, complementing the goal of APT, thereby enhancing context learning and helping the model better understand the task. In contrast, while the entity-word MDT somewhat alleviates the few-shot distillation dilemma, its lack of contextual information offers limited assistance in helping the model understand the task.
\subsection{Effectiveness}
The effectiveness of our proposed method is reflected in two aspects: \par 1)\textbf{ No additional data required during training.} Unlike the \cite{wang2022few}, our method does not rely on additional synthetic data during the training process. \cite{wang2022few} requires a complex and time-consuming data synthesis process, along with carefully designed adversarial matching to ensure data validity and authenticity. In contrast, our method avoids catastrophic forgetting by adding MDT, allowing the model to recall previous knowledge effectively. \par 2) \textbf{Only one decoding pass needed during evaluation.} Traditional prompt-based methods often require enumerating different spans of entity mentions during evaluation, which is not only time-consuming but also causes decoding time to increase with sequence length. In contrast, our method requires only one decoding pass during evaluation.
\section{Conclusion}

we address specific challenges faced in few-shot continual learning named entity recognition by proposing a straightforward and efficient solution. By integrating anchor words-oriented prompt tuning with memory demonstration templates, our approach not only avoids the few-shot distillation dilemma but also enhances the model's generalization and adaptability in dynamic data streams.

\clearpage
\bibliography{custom}           

\appendix

\section{Implementation Details}
\label{sec:appendix}

We use bert-base-cased as the PLM, with a hidden layer size of 768. All parameters are fine-tuned using the Adam \cite{kingma2014adam} optimizer, with the learning rate for the BERT encoder set to 1e-4. For the base class stage, training is conducted for 5 epochs with a batch size of 32, and no BERT parameters are frozen. For the few-shot incremental stages, training is conducted for 20 epochs with a batch size of 2, and the first 9 layers of BERT are frozen. Since we do not have a validation set, the model from the last epoch of training is used for final inference. The number of memory demonstration templates for each class is set to 2. All results in this paper use the Macro-averaged F1 Score as the final evaluation metric. All training was performed on an NVIDIA RTX 4090 GPU with 24GB of memory.




\section{ Dataset Statistics}
We have listed the detailed statistics of the two original datasets used in our study in \autoref{table4}. We utilized 4 entity types on CoNLL2003 and 18 entity types on OntoNote 5.0. To align our experiments with real-world few-shot scenarios, we did not set up a validation set. Instead, the training set for the incremental phases was derived from few-shot samples on the original validation set.

\begin{table}[h!]
\centering
\caption{Statistics of Two Datasets}
\begin{tabular}{ccccc}
\hline
\multirow{2}{*}{\textbf{Datasets}} & \multicolumn{3}{c}{$|{\cal D}|$} & \multirow{2}{*}{\textbf{\# Types}} \\ \cline{2-4}
 & \textbf{Train} & \textbf{Val} & \textbf{Test} &  \\ \hline
\textbf{CoNLL2003} & 14,987 & 3,466 & 3,684 & 4 \\ \hline
\textbf{OntoNote5.0} & 59,924 & 8,528 & 8,262 & 18 \\ \hline
\end{tabular}
\label{table4}
\end{table}

\section{Reorganization Strategies} 
Unlike the \cite{zhang2023neural}, which divides the training set into Split and Filtered, our training set does not involve a Split setting. Instead, we reorganize the base class training set into Train on All (ToA) and Train on Filtered (ToF). For the evaluation set, we reorganize it into Evaluate on All (EoA) and Evaluate on Filtered (EoF). We conducted experiments under the following four combinations to comprehensively evaluate our proposed method:
\par $ToA \to EoA$ : Training on all available training data and evaluating on all test data. This is the standard reorganization strategy that is consistent with most baselines.
\par $ToA \to EoF$ : Training on all available training data and evaluating only on test data related to tasks encountered so far.
\par$ToF \to EoA$ : Training only on data related to the current task and evaluating on all test data.
\par $ToF \to EoF$ : Training only on data related to the current task and evaluating only on test data related to tasks encountered so far.

\begin{table*}[h]
\centering
\begin{small}

\caption{Different Task Permutations on Two Datasets.}
\scalebox{1}{
\begin{tabular}{cl}
\hline
Datasets   & \multicolumn{1}{c}{Permutations} \\ \hline
           & ${P_1}$: \{PER\} $\Rightarrow$ \{LOC\} $\Rightarrow$ \{ORG\} $\Rightarrow$ \{MISC\} \\
           & ${P_2}$: \{PER\} $\Rightarrow$ \{MISC\} $\Rightarrow$ \{LOC\} $\Rightarrow$ \{ORG\} \\
           & ${P_3}$: \{LOC\} $\Rightarrow$ \{PER\} $\Rightarrow$ \{ORG\} $\Rightarrow$ \{MISC\} \\
           & ${P_4}$: \{LOC\} $\Rightarrow$ \{ORG\} $\Rightarrow$ \{MISC\} $\Rightarrow$ \{PER\} \\
CoNLL2003  & ${P_5}$: \{ORG\} $\Rightarrow$ \{LOC\} $\Rightarrow$ \{MISC\} $\Rightarrow$ \{PER\} \\
           & ${P_6}$: \{ORG\} $\Rightarrow$ \{MISC\} $\Rightarrow$ \{PER\} $\Rightarrow$ \{LOC\} \\
           & ${P_7}$: \{MISC\} $\Rightarrow$ \{PER\} $\Rightarrow$ \{LOC\} $\Rightarrow$ \{ORG\} \\
           & ${P_8}$: \{MISC\} $\Rightarrow$ \{ORG\} $\Rightarrow$ \{PER\} $\Rightarrow$ \{LOC\} \\ \hline
           & \begin{tabular}[c]{@{}l@{}}${P_1}$: \{CARDINAL, DATE, EVENT, FAC\} $\Rightarrow$ \{GPE, LANGUAGE\} \\ $\Rightarrow$ \{LAW\} \\ $\Rightarrow$ \{LOC, MONEY\} $\Rightarrow$ \{NORP\} \\ $\Rightarrow$ \{ORDINAL, ORG\} \\ $\Rightarrow$ \{PERCENT\} $\Rightarrow$ \{PERSON, PRODUCT\} \\ $\Rightarrow$ \{QUANTITY, TIME, WORK\_OF\_ART\}\end{tabular} \\
OntoNote5.0 & \\
           & \begin{tabular}[c]{@{}l@{}}${P_2}$: \{CARDINAL, DATE, EVENT, FAC\} $\Rightarrow$ \{GPE\} \\ $\Rightarrow$ \{LANGUAGE\} $\Rightarrow$ \{LAW\} \\ $\Rightarrow$ \{LOC\} $\Rightarrow$ \{MONEY, NORP\} \\ $\Rightarrow$ \{ORDINAL, ORG\} \\ $\Rightarrow$ \{PERCENT, PERSON\} \\ $\Rightarrow$ \{PRODUCT, QUANTITY\} \\ $\Rightarrow$ \{TIME, WORK\_OF\_ART\}\end{tabular} \\ \hline
\end{tabular}
}
\label{table5}
\end{small}
\end{table*}

\begin{table*}[]
\centering
\begin{small}
\caption{Representative entity words used in our experiments.}
\begin{tabular}{ll}
\hline
\textbf{Datasets} & \textbf{Representative Entity Words} \\ \hline
 & \{ \\
 & "A-PER": {[}"Michael", "John", "David", "Thomas", "Martin", "Paul"{]}, \\
 & "A-ORG": {[}"Corp", "Inc", "Commission", "Union", "Bank", "Party"{]}, \\
\textbf{CoNLL2003} & "A-LOC": {[}"England", "Germany", "Australia", "France", "Russia", "Italy"{]}, \\
 & "A-MISC": {[}"Palestinians", "Russian", "Chinese", "Dutch", "Russians", "English"{]} \\
 & \} \\ \hline
 & \{ \\
 & "A-CARDINAL": {[}"one", "two", "three", "four", "five", "six"{]}, \\
 & "A-DATE": {[}"today", "yesterday", "September", "Monday", "Friday", "Today"{]}, \\
 & "A-EVENT": {[}"War", "Games", "Katrina", "Year", "Hurricane", "II"{]}, \\
 & "A-FAC": {[}"Airport", "Bridge", "Base", "Memorial", "Canal", "Guantanamo"{]}, \\
 & "A-GPE": {[}"US", "China", "United", "Beijing", "Israel", "Taiwan"{]}, \\
 & "A-LANGUAGE": {[}"Mandarin", "Streetspeak", "Romance", "Ogilvyspeak", "Pentagonese", "Pilipino"{]}, \\
 & "A-LAW": {[}"Chapter", "Constitution", "Code", "Amendment", "Protocol", "RICO"{]}, \\
 & "A-LOC": {[}"Middle", "River", "Sea", "Ocean", "Mars", "Mountains"{]}, \\
\textbf{OntoNote5.0} & "A-MONEY": {[}"billion", "million", "\$"{]}, \\
 & "A-NORP": {[}"Chinese", "Israeli", "Palestinians", "American", "Japanese", "Palestinian"{]}, \\
 & "A-ORDINAL": {[}"first", "second", "third", "First", "fourth", "eighth"{]}, \\
 & "A-ORG": {[}"National", "Corp", "News", "Inc", "Senate", "Court"{]}, \\
 & "A-PERCENT": {[}"\%"{]}, \\
 & "A-PERSON": {[}"John", "David", "Peter", "Michael", "Robert", "James"{]}, \\
 & "A-PRODUCT": {[}"USS", "Discovery", "Cole", "Atlantis", "Coke", "Galileo"{]}, \\
 & "A-QUANTITY": {[}"gallon", "miles", "degrees", "ton", "meter", "degrees"{]}, \\
 & "A-TIME": {[}"tonight", "night", "morning", "evening", "afternoon", "hours"{]}, \\
 & "A-WORK\_OF\_ART": {[}"Prize", "Nobel", "Late", "Morning", "PhD", "Edition"{]} \\
 & \} \\ \hline
\end{tabular}
\end{small}
\end{table*}

\section{CL Task Permutations} 
Table 4 presents the different task permutations on the two datasets, which strictly follow the settings from \cite{wang2022few} to ensure a fair comparison.

\section{Selection of Entity Words} 

\autoref{table5} shows the representative entities for each category, most of which were selected using the Data\&LM+Virtual method, with a few selected based on class names and high-frequency words from the dataset.

\end{document}